\documentclass[conference]{IEEEtran}
\usepackage{times}

\usepackage[numbers]{natbib}
\usepackage{multicol}
\usepackage[bookmarks=true]{hyperref}
\usepackage{amsfonts}
\usepackage{amsmath}
\usepackage{comment}
\usepackage{graphicx}
\usepackage{wrapfig}

\usepackage[ruled,vlined]{algorithm2e}
\usepackage{xcolor}

\definecolor{commentcolor}{RGB}{200, 120, 0}

\usepackage{tcolorbox}
\tcbuselibrary{breakable}

\usepackage{hyperref}

\usepackage{ifthen,version}
\usepackage{color}
\newboolean{include-notes}
\setboolean{include-notes}{false}
\newcommand{\todo}[1]{\ifthenelse{\boolean{include-notes}}%
{\textcolor{red}{\textbf{TODO: #1}}}{}}

\newcommand{\method}{ASQ} 

\newcommand{\hmnote}[1]{\ifthenelse{\boolean{include-notes}}%
{\textcolor{blue}{\textbf{HM: #1}}}{}}
\newcommand{\nwnote}[1]{\ifthenelse{\boolean{include-notes}}%
{\textcolor{red}{\textbf{N: #1}}}{}}
\newcommand{\abnote}[1]{\ifthenelse{\boolean{include-notes}}%
{\textcolor{purple}{\textbf{AB: #1}}}{}}

\begin{document}

\newtcolorbox{promptbox}{colback=gray!8, colframe=gray!30, 
  fonttitle=\bfseries, title=Prompt, left=4pt, right=4pt, breakable, coltitle=black}

\title{Robots That Know What to Ask: Recovering Misaligned Rewards through Targeted Explanations}

\author{
\authorblockN{Helena Merker, Nick Walker, Andreea Bobu}
\authorblockA{Massachusetts Institute of Technology \\
\{hmerker, nswalker, abobu\}@mit.edu}
}

\maketitle

\begin{abstract}

    Learning reward functions from demonstrations assumes that demonstrations provide adequate supervision over all \textit{features}—or task-relevant aspects of behavior. In practice, demonstrations are often imperfect: humans may under-emphasize certain features due to cognitive load or physical difficulty, or the training regime may fail to sufficiently cover all relevant situations. In either case, important features may be underspecified, leading to ambiguity in the learned reward function and misaligned behavior at deployment. We propose a framework that detects such underspecified features and actively solicits targeted corrective demonstrations. Our key insight is that demonstrations implicitly reveal which features are well specified: features that are consistently optimized show little variation across demonstrations, while features that are underspecified vary widely. We leverage this statistical signal to infer which features may have been insufficiently demonstrated. The robot then explains which features it is uncertain about in natural language and queries for demonstrations that explicitly address the identified gaps. We evaluate our approach in a simulated tabletop manipulation domain and in a user study with a real Franka robot. Targeted, explanation-guided queries significantly improve reward recovery compared to random querying and passive data collection, reducing ambiguity that would otherwise persist in learning from imperfect demonstrations.

\end{abstract}

\IEEEpeerreviewmaketitle

\section{Introduction}

Imagine teaching a robot how to carry a coffee cup across a cluttered table. As you demonstrate the task, you focus on keeping the cup upright and avoiding collisions, but your trajectory passes at varying distances from a nearby laptop. Although maintaining a safe distance from the laptop \textit{is} important, consistently controlling cup orientation, collision avoidance, and object proximity all at once is difficult. When the robot is later deployed, it reliably keeps the cup upright but passes uncomfortably close to the laptop. From the demonstrations alone, the robot cannot tell whether proximity to the laptop was unimportant, or whether the demonstrations simply failed to convey how that feature should be handled.

This illustrates a fundamental challenge in learning from demonstrations: observed behavior alone cannot distinguish whether a feature varies because the user is genuinely indifferent, or simply because they didn't emphasize it sufficiently. Such underspecification can happen for many reasons. Demonstrating a task is cognitively demanding~\cite{amershi2019guidelines}, and humans may not always attend to every relevant feature with equal care~\cite{bajcsy2018onefeature}. Even when they try, the training regime itself may lack sufficient coverage~\cite{peng2023DFA}: certain features may remain constant, rare, or difficult to exercise (e.g., obstacles positioned at the edges of the workspace), making it impossible for demonstrators to convey how they should be handled. In both cases, demonstrations underspecify parts of the reward function, creating ambiguity that cannot be resolved from the data alone.

\begin{figure}
    \centering
    \includegraphics[width=\linewidth]{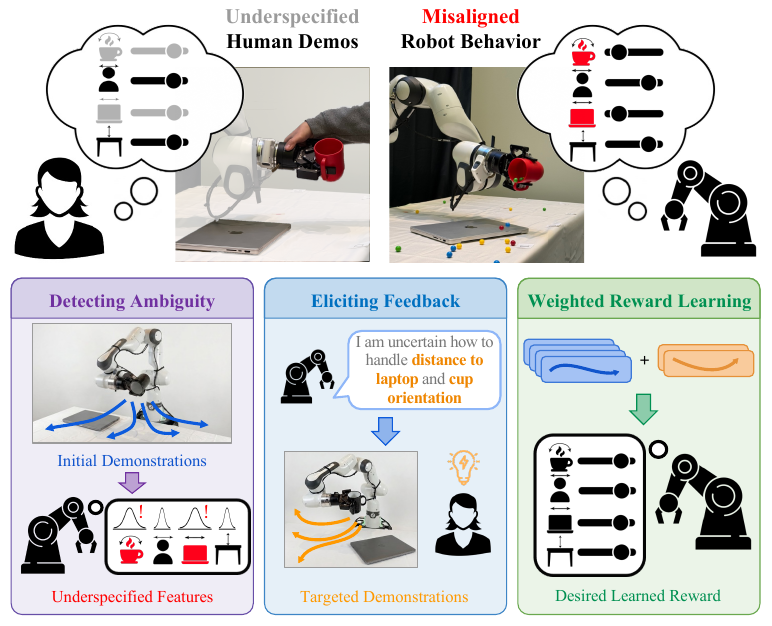}
    \caption{Humans don't or can't always attend to all important task features when providing demonstrations (top left), which can lead to a learner inferring a misaligned reward (top right). Our method resolves this challenge by identifying when features are likely underspecified and asking for targeted, explanation-guided demonstrations (bottom).}
    \label{fig:hero}
\end{figure}

Despite this ambiguity, current approaches implicitly assume that demonstrations provide complete supervision over all task-relevant features, and that any behavior which is inconsistent across demonstrations is evidence of low reward weight~\cite{ziebart2008maximum,finn2016guided}.
When that assumption fails, the learned objective may not capture all the considerations that a human expects the robot to account for, leading to suboptimal or unsafe behavior at deployment (e.g., a robot that optimizes for efficiency while ignoring proximity to fragile objects or people)~\cite{bobu2020quant}. 

 In this work, we introduce Ambiguity-Sensitive Querying for Reward Learning (\method), a framework that resolves this reward ambiguity by enabling the robot to diagnose which features are underspecified and to solicit targeted corrective demonstrations. Our key observation is that optimization leaves a statistical footprint in the variability of feature values across demonstrations. Features that demonstrators actively optimize tend to show low variability across demonstrations, whereas underspecified features tend to vary widely. By modeling the distribution of feature variability under attended and unattended conditions, we can identify which features were likely underspecified during demonstration.

Since these features may have been deliberately ignored, our method queries users to distinguish underspecification from genuine indifference. 
The robot provides a natural language explanation that identifies the features that it is uncertain about, directing the demonstrator's attention to the specific dimensions where further supervision may be needed. We then instruct the user to provide additional demonstrations while explicitly considering how the robot should handle the identified features. By directing the user's attention to specific features, the resulting demonstrations vary in ways that reflect how those features should influence behavior, resolving ambiguity in the learned reward.
We empirically validate our framework across a simulated continuous 7DoF robotic manipulation task, as well as in a user study with a physical Franka robot. We show that targeted queries recover accurate reward functions, and that real users provide more informative corrective demonstrations when given feature-based explanations compared to no guidance or simply observing robot behavior.

\section{Related Work}

\subsection{Learning from Human Input}

Inferring a policy or reward function from human inputs, such as demonstrations \cite{ziebart2008maximum,ng2000irl,finn2016guided}, comparisons \cite{christianoLBMLA17}, corrections \cite{bajcsy2017learning,bobu2018learning}, and trajectory rankings \cite{brownCSN20}, has proven effective for learning for robot tasks~\cite{ravichandar2020lfd}. Inverse reinforcement learning methods typically rely upon a set of predefined features that capture task-relevant aspects of the environment \cite{abbeel2004apprenticeship}. Prior work has explored various forms of corrective human feedback, including natural language \cite{sharmaSBPH0AF22}, physical interaction \cite{bajcsy2018onefeature}, and teleoperation \cite{rakitaMG18}. Even when task-relevant features are available, users may not cleanly separate the intended signal for each feature, making it difficult to determine how each feature should influence behavior \cite{bajcsy2018onefeature}. ASQ detects and resolves such ambiguity by identifying which features received inadequate supervision and eliciting targeted feedback.

\subsection{Learning from Incomplete or Imperfect Demonstrations} %

Inverse reinforcement learning methods often assume that the demonstrator is at least noisy-rational with respect to some underlying reward~\cite{ziebart2008maximum}. In practice, demonstrations can be inconsistent or lack coverage of the state space ~\cite{cakmak2010optimality,sakr2025consistency,belkhale2023quality}. Prior work attempts to address this issue by collecting additional feedback modalities, such as pairwise preferences~\cite{brown2019drex} or natural language instructions~\cite{hwang2025maskedirlllmguidedreward}, modeling demonstrator suboptimality~\cite{beliaev2025estimating}, or explicitly quantifying objective misspecification~\cite{bobu2020quant}. These approaches improve learning from fixed data but do not actively diagnose which features are underspecified. \citet{peng2023DFA} come closest, proposing a human-in-the-loop framework that diagnoses policy failures and solicits corrections, but at the level of state-space coverage rather than feature supervision.
ASQ detects sub-optimality and actively queries for additional demonstrations targeting specific features, creating heterogeneous demonstrations that we can weight distinctly to achieve more efficient learning.

\subsection{Queries and Explanations in Human-Robot Teaching} %

Prior work has characterized the types of queries a learner can pose, including label, demonstration, and feature queries, along with how naturally humans answer each~\cite{cakmak2012designing}. Algorithmic approaches actively select queries that reduce reward uncertainty, such as pairwise trajectory comparisons~\cite{sadigh2017active}, or choose informative behaviors that help users build a mental model of the robot's objective~\cite{huang2019enabling}. A complementary line of work focuses on communicating what the robot has already learned to the user: counterfactual demonstrations help users isolate task-relevant state concepts~\cite{peng2023DFA}, sampled policy rollouts explain the inferred reward~\cite{gu2025xai}, and expressive motion can convey both the robot's intent and its inability to succeed~\cite{kwon2018incapability}, across visual, haptic, and auditory channels~\cite{habibian2025communicating}. ASQ identifies features that need additional supervision, which may be used as a basis to generate queries expressed in any modality.

\section{Problem Definition}

We address the challenge of inferring a human's latent preferences from a limited set of demonstrations. We formalize the environment as a Markov Decision Process (MDP) defined by the tuple $\mathcal{M} = \langle S, A, T, R \rangle$, where $S$ is the state space, $A$ is the action space, $T$ is the transition dynamics, and $R$ is the reward function. The MDP is solved by computing a policy $\pi: S \rightarrow A$, a mapping that prescribes which action the robot should execute in each state. Under the inverse reinforcement learning (IRL) formulation, we assume the robot does not know the reward function and must instead attempt to infer it from the human's input about the task (e.g., demonstrations). A key challenge is that demonstrations (especially when given by non-experts) may be imperfect: important aspects of the task may receive insufficient attention due to cognitive constraints, physical difficulty, or limited coverage in the training regime.

\smallskip
\noindent\textbf{Reward Representation.} We adopt a linear reward parameterization over state features. Let $\phi: S \rightarrow \mathbb{R}^k$ denote a feature mapping from states to a $k$-dimensional feature vector, with $\theta \in \mathbb{R}^k$ representing the reward weights. The reward at a given state is $R(s) = \theta^\mathsf{T} \phi(s)$. A trajectory $\tau = (s^0, ..., s^T)$ belongs to the space of all trajectories $\Xi$. The cumulative reward for a trajectory is obtained by aggregating over states:
\begin{equation}
R(\tau) = \sum_{s^t \in \tau} \theta^\mathsf{T} \phi(s^t) = \theta^\mathsf{T} \phi(\tau)
\end{equation}
where $\phi(\tau) = \sum_{s^t \in \tau} \phi(s^t)$ accumulates features along the trajectory. We denote the human's true preferences as $\theta^* \in \mathbb{R}^k$.

The choice of features $\phi$ is critical to the expressiveness of learnable reward functions. In practice, these features can be hand-crafted~\cite{ziebart2008maximum}, learned \cite{bobu2022inducing}, or even LLM-generated~\cite{peng24algae}. Recent progress in LLMs in particular has made it possible to easily extract task-relevant features from high-level descriptions of the task. Throughout this work, we assume access to such a candidate set $\phi = \{\phi_i\}^{k}_{i=1}$, and we study how uneven supervision over these features affects reward inference.

\smallskip
\noindent\textbf{Modeling Demonstrations.} We model the demonstrations as arising from a Boltzmann noisily-rational process~\cite{jaynes1957infotheory,von1945theory,baker2007goal}, where the human tends to choose trajectories in proportion to their exponentiated reward. In classic formulations~\cite{fisac2018probabilistically,bobu2018learning,bobu2020quant}, a single inverse-temperature parameter $\beta$ captures how rationally the human optimizes their intended reward function: $P(\tau \mid \theta^{*}, \beta) \propto \exp\left(\beta \cdot {\theta^*}^T \phi(\tau) \right)$. This assumes uniform rationality across all features, which may not hold when demonstrators have limited attention, face varying task difficulty, or when the training regime provides insufficient coverage to adequately exercise certain features. We generalize this to a per-feature rationality vector
$\beta \in \mathbb{R}^k_{\ge 0}$, where each component $\beta_{i}$ modulates how consistently the demonstrations attend to feature $\phi_i$. The probability of generating a trajectory $\tau$ is given by:
\begin{equation}
P(\tau \mid \theta^{*}, \beta) \propto \exp\left(\sum_{i=1}^k \beta_i \theta^{*}_i \phi_i(\tau) \right)
\label{eq:obs_model}
\end{equation}
Here, $\beta_i$ represents the human's attention or ``care'' for feature $\phi_i$: higher values indicate the human more reliably produces behavior consistent with that feature, while lower values indicate greater variability--the demonstrations provide weaker signal about that feature's role in the reward.

\smallskip
\noindent\textbf{The Underspecification Problem.} Demonstrations may provide uneven supervision across different aspects of a task. Formally, this occurs when the rationality vector $\beta$ has $\beta_i \approx 0$ for one or more features, meaning that certain features receive insufficient attention during demonstration. 
Such underspecification can arise when the demonstrated behavior involves several objectives, when the user is demonstrating under increased cognitive load, or when the configuration of the training environment makes it difficult to exercise a particular feature. This uneven supervision creates a fundamental identifiability problem. When $\beta_i \approx 0$, the demonstrations are uninformative about $\theta_i^*$: any value of the true weight produces the same likelihood for the observed data. The demonstrations alone cannot disambiguate whether a feature was deliberately deprioritized by the human ($\theta_i^* \approx 0$) or simply not attended to during the demonstration process ($\beta_i \approx 0$). 

Given initial demonstrations $D_\mathrm{init} \sim P(\tau \mid \theta^*, \beta)$, our goal is to identify which features received insufficient supervision, then elicit targeted corrective demonstrations that specifically address these features. We generate natural language explanations that make the robot's inferred preferences explicit.
These explanations prompt demonstrators to provide corrective demonstrations $D_\mathrm{extra}$ that address the underspecified features, enabling recovery of the true weights $\theta^*$ through targeted human feedback.

\section{Method}

Our approach detects features for which the demonstrations provide weak supervision, then queries the human to clarify their preferences over these features. %
We leverage a key insight: features that are actively optimized exhibit low variance across demonstrations, while underspecified features vary widely. 
We formalize this intuition through Bayesian model selection given observed variances, calibrated against reference distributions of what variance looks like under different optimization conditions.
Having identified likely underspecified features, we query the demonstrator about them using natural language explanations. %
This targeted intervention enables us to distinguish whether high variance reflected genuine indifference to a feature or merely incidental underspecification during the initial demonstrations. We then combine the original and targeted demonstrations, weighting each set according to which features received attention, to recover reward weights that reflect the human's complete preferences. We present the full procedure in \hyperref[alg:asqrl]{Algorithm~\ref*{alg:asqrl}}.

\begin{algorithm}[t]
\caption{Ambiguity-Sensitive Querying (\method)}
\label{alg:asqrl}
\KwIn{Initial demonstrations $D_\mathrm{init}$, task-relevant features $\phi$, pre-trained reference distributions $P(\sigma^2_i \mid o_i, \phi_i)$ for $\phi_i \in \phi$, $o_i \in \{0,1\}$}

\For{$\phi_i \in \phi$}{
    $\sigma^2_i \gets \mathrm{Var}(\{\phi_i(\tau) : \tau \in D_\mathrm{init}\})$
}

\For{$\phi_{\mathrm{hyp}} \subseteq \phi$}{
    $P(\phi_{\mathrm{hyp}} \mid \{\sigma^2_i\}) \gets \mathrm{ComputePosterior}(\phi_{\mathrm{hyp}}, \{\sigma^2_i\}, \{P(\sigma^2_i \mid o_i, \phi_i)\})$
}

$\phi_{\mathrm{under}} \gets \arg\max_{\phi_{\mathrm{hyp}}} P(\phi_{\mathrm{hyp}} \mid \{\sigma^2_i\})$

$\texttt{query} \gets \mathrm{GenerateExplanation}(\phi_{\mathrm{under}})$

$D_\mathrm{extra} \gets \mathrm{QueryDemonstrator}(\texttt{query})$

$D_\mathrm{total} \gets D_\mathrm{init} \cup D_\mathrm{extra}$

$\beta^{\mathrm{init}}, \beta^{\mathrm{extra}} \gets \mathrm{AssignRationalityWeights}(\phi_{\mathrm{under}}, \phi)$

$\theta^* \gets \mathrm{WeightedMaxEntIRL}(D_\mathrm{total}, \beta^{\mathrm{init}}, \beta^{\mathrm{extra}})$

\KwOut{Learned reward weights $\theta^*$}
\end{algorithm}

\subsection{Detecting Underspecified Features}

We begin by identifying features for which the demonstrations provide weak supervision. Under our demonstration model, feature-level optimization leaves a statistical footprint in the variability of observed feature values across demonstrations. Features that are consistently optimized tend to exhibit low variance, while features that receive little attention vary more widely.
Concretely, if a demonstrator reliably attends to keeping a robot far from an obstacle, the corresponding feature values will cluster tightly across trajectories. Conversely, if obstacle distance was not a focus---either because the human deemed it unimportant ($\theta_i^*$ low) or simply did not think to vary it informatively ($\beta_i$ low)---we expect to see greater spread in the observed feature values. 
We treat high variance as a signal that a feature may have been underspecified and therefore requires clarification from the user.

\smallskip
\noindent\textbf{Inferring Optimization Status.}
Our goal is to identify which features were not consistently optimized in the demonstrations and are therefore potentially underspecified. We treat this as a latent inference problem: given observed feature variability, we infer which features the demonstrator did not reliably attend to. To formalize this notion, we introduce a binary indicator $o_i \in \{0, 1\}$ for each feature $\phi_i$, where $o_i = 1$ denotes that feature $\phi_i$ was consistently optimized in the demonstrations and $o_i = 0$ denotes otherwise. Throughout this section, we use the terms ``non-optimized'' and ``underspecified'' interchangeably to refer to features with $o_i = 0$.

Given initial demonstrations $D_\mathrm{init} = \{\tau_1, \ldots, \tau_n\}$ and the feature set $\phi = \{\phi_i\}^{k}_{i=1}$, we compute the empirical variance for each feature across demonstrations:
$\sigma^2_i = \frac{1}{n-1} \sum_{j=1}^n (\phi_i(\tau_j) - \bar{\phi}_i)^2$,
where $\bar{\phi}_i = \frac{1}{n}\sum_{j=1}^n \phi_i(\tau_j)$ is the sample mean. As discussed above, high variance indicates that a feature may not have been consistently optimized, while low variance suggests reliable attention during demonstration.

We frame detection as a Bayesian model selection problem. Each hypothesis $\phi_{\mathrm{hyp}} \subseteq \phi$ represents a candidate set of underspecified features. Given the observed variances, we compute the posterior probability of each hypothesis:
\begin{equation}
P(\phi_{\mathrm{hyp}} \mid \{\sigma^2_i\}) \propto P(\{\sigma^2_i\} \mid \phi_{\mathrm{hyp}}) \cdot P(\phi_{\mathrm{hyp}})\enspace.
\end{equation}
The likelihood $P(\{\sigma^2_i\} \mid \phi_{\mathrm{hyp}})$ captures how well the observed variance pattern matches what we would expect if exactly the features in $\phi_{\mathrm{hyp}}$ were non-optimized. Assuming conditional independence of feature variances given the optimization status, the likelihood factorizes as:
\begin{equation}
P(\{\sigma^2_i\} \mid \phi_{\mathrm{hyp}}) = \prod_{i \in \phi} P(\sigma^2_i \mid o_i, \phi_i)\enspace,
\end{equation}
where $o_i = 0$ if $\phi_i \in \phi_{\mathrm{hyp}}$ and $o_i = 1$ otherwise. The individual likelihoods $P(\sigma^2_i \mid o_i, \phi_i)$ are obtained from reference distributions that characterize feature variance under optimized and non-optimized conditions; we describe how we construct these distributions below. 
We select the set of likely underspecified features $\phi_{\mathrm{under}}$ via maximum a posteriori inference: $\phi_{\mathrm{under}} = \arg\max_{\phi_{\mathrm{hyp}}} P(\phi_{\mathrm{hyp}} \mid \{\sigma^2_i\})$.

\smallskip
\noindent\textbf{Filtering to Task-Relevant Features.} Before analyzing feature variability, we restrict the robot's attention to a candidate set of task-relevant features. Irrelevant features---those with $\theta^*_i \approx 0$ regardless of demonstrator attention---will naturally exhibit high variance, but querying about them wastes the human's limited attention. Presenting a user with many candidate features when only a few actually matter imposes unnecessary cognitive load and degrades feedback quality~\cite{amershi2019guidelines,sweller1988cognitive}. By filtering to task-relevant features, we focus the clarification interaction on dimensions where human input will be most valuable. Moreover, Bayesian model selection over feature subsets scales exponentially in $|\phi|$, making it essential to limit the hypothesis space to a small, task-relevant set.

In practice, the task-relevant feature set can be identified using domain knowledge or inferred from an LLM that rates feature relevance given a task description~\cite{peng24plga}. In this work, we use an LLM to provide a prior over feature subsets, effectively approximating $P(\phi_{\mathrm{hyp}})$. This leverages the fact that LLMs encode rich common-sense priors about task structure and human-relevant considerations~\cite{bubek2023sparks}. By concentrating probability mass on hypotheses involving features a human is likely to care about, this prior reduces false positives and focuses clarification on meaningful dimensions.

\smallskip
\noindent\textbf{Constructing Reference Distributions.} The raw variance $\sigma_i^2$ is difficult to interpret in isolation: what constitutes ``high'' variance for one feature may be typical for another, depending on the feature's natural scale, environment dynamics, and inherent task variability. To calibrate our expectations, we construct reference distributions that characterize variance under different optimization conditions.

Specifically, for each feature $\phi_i \in \phi$, we seek to estimate $P(\sigma_i^2 \mid o_i=1,\phi_i)$ and $P(\sigma_i^2 \mid o_i=0,\phi_i)$. We build these distributions by simulating trajectories under policies optimizing different rewards $R_{\theta}$ to ensure diverse human intents. For each policy, some features are assigned positive weight (and thus optimized) while others are assigned zero weight (and thus ignored). By aggregating variance observations across policies that include $\phi_i$ in their objective versus those that exclude it, we obtain empirical samples from each reference distribution. Although individual feature values are not normally distributed, we empirically find that their sample variances are approximately Chi-squared distributed, which facilitates subsequent likelihood computation.

\subsection{Eliciting Targeted Feedback}

Having identified features for which the demonstrations provide weak supervision, we query the demonstrator to resolve the ambiguity. This query constitutes a causal intervention: by explicitly directing the user's attention to specific features, we can distinguish whether low optimization reflected genuine indifference or incidental underspecification. Specifically, we generate a natural language explanation that communicates which features the robot is uncertain about. The explanation follows a fixed template that describes the underspecified features in terms the demonstrator can act upon (e.g., ``I am uncertain about how to handle distance to the laptop''). After presenting this explanation, we request additional demonstrations with the instruction to perform the task while considering how the robot should handle the identified features. We then collect $M$ additional demonstrations, each of which attends to the previously underspecified features $\phi_{\mathrm{under}}$.

\subsection{Reward Learning}

Standard IRL typically assumes that all demonstrations are generated under a single observation model, treating them as equally informative when inferring the reward weights~\cite{ziebart2008maximum,finn2016guided}. Under this formulation, we would naively pool initial and targeted demonstrations into a single dataset $D_\mathrm{total} = D_\mathrm{init} \cup D_\mathrm{extra}$ and 
recover $\theta^{*}$ by maximizing their log-likelihood:
\begin{equation}
\theta^{*} \;=\; \arg\max_{\theta}\; \sum_{\tau \in D_\mathrm{total}} \log P(\tau \mid \theta, \beta)\enspace,
\label{eq:ll}
\end{equation}
where $P(\tau \mid \theta, \beta)$ could be instantiated as the Boltzmann noisily-rational model in Eq.~\ref{eq:obs_model}, with either a fixed rationality parameter~\cite{liu2016goal} or one jointly estimated with $\theta$~\cite{bobu2020quant}.

This formulation implicitly assumes that all demonstrations reflect the same pattern of attention across features---i.e., that they are generated using the same rationality vector $\beta$. In our setting, however, this assumption is violated by construction. The initial demonstrations $D_{\mathrm{init}}$ are collected without explicit guidance, and may therefore reflect uneven attention across features, while the targeted demonstrations $D_{\mathrm{extra}}$ are collected after \textit{explicitly} directing the demonstrator’s attention to a subset of previously underspecified features. As a result, the two demonstration sets arise from different attention patterns and should not be modeled using the same rationality vector.

To capture this distinction, we model the initial and targeted demonstrations using separate rationality vectors, $\beta^{\mathrm{init}}$ and $\beta^{\mathrm{extra}}$, while assuming a shared underlying reward function parameterization $\theta^*$. Intuitively, $\beta^{\mathrm{init}}$ reflects the features the demonstrator naturally attended to during unguided demonstration, whereas $\beta^{\mathrm{extra}}$ reflects the features they were explicitly instructed to attend to during targeted feedback. 

Since we explicitly asked the demonstrator to attend to the features in $\phi_{\mathrm{under}}$, we assign $\beta^{\mathrm{extra}}_i = \beta_{\mathrm{high}}$ for $\phi_i \in \phi_{\mathrm{under}}$, and low rationality $\beta_{\mathrm{low}}$ for the remaining features. Conversely, for the initial demonstrations, we reflect their inferred lack of attention by setting $\beta^{\mathrm{init}}_i = \beta_{\mathrm{low}}$ for $\phi_i \in \phi_{\mathrm{under}}$ and $\beta^{\mathrm{init}}_i = \beta_{\mathrm{high}}$ for all other features. In our implementation, we estimate $\beta_{\mathrm{low}}$ and $\beta_{\mathrm{high}}$ via hyperparameter tuning on a held-out validation set with simulated human feedback.

To solve Eq.~\ref{eq:ll}, one can minimize the negative log-likelihood with an off-the-shelf optimizer. Under standard IRL, the negative log-likelihood loss for a single demonstration set $D$ is
\begin{equation}
\mathcal{L}(D;\theta,\beta) = \frac{1}{|D|} \sum_{\tau \in D} \left[-\sum_{i=1}^k \beta_i \theta_i \phi_i(\tau)\right] + \log Z(\theta,\beta) \enspace,
\end{equation}
where $Z$ is the partition function that normalizes the observation model in Eq.~\ref{eq:obs_model}.

In our setting, the initial and targeted demonstrations are generated under different attention regimes and are therefore modeled with different rationality vectors.
We therefore define the loss as the sum of the negative log-likelihoods of the two demonstration sets: 
\begin{equation}
\mathcal{L}_{\mathrm{total}} = \frac{\lvert D_{\mathrm{init}} \rvert \cdot \mathcal{L}_{\mathrm{init}} + \lvert D_{\mathrm{extra}} \rvert \cdot \mathcal{L}_{\mathrm{extra}}}{\lvert D_{\mathrm{init}} \rvert + \lvert D_{\mathrm{extra}} \rvert}
\end{equation}
\begin{equation}
\mathcal{L}_{\mathrm{init}} = \frac{1}{|D_{\mathrm{init}}|} \sum_{\tau \in D_{\mathrm{init}}} -\sum_{i=1}^{k} \beta^{\mathrm{init}}_i \theta_i \phi_i(\tau) + \log Z(\theta, \beta^{\mathrm{init}})
\end{equation}
\begin{equation}
\mathcal{L}_{\mathrm{extra}} = \frac{1}{|D_{\mathrm{extra}}|} \sum_{\tau \in D_{\mathrm{extra}}} -\sum_{i=1}^{k} \beta^{\mathrm{extra}}_i \theta_i \phi_i(\tau) + \log Z(\theta, \beta^{\mathrm{extra}})
\end{equation}
where $Z(\theta,\beta) = \int \exp\left(\sum_{i=1}^{k} \beta_i \theta_i \phi_i(\tau')\right) d\tau'$ is approximated via importance sampling as in prior work~\cite{finn2016guided}. 
The combined demonstration set $D_\mathrm{total}$ contains informative signal about all task-relevant features. The initial demonstrations $D_\mathrm{init}$ provide supervision for features the user naturally attended to during demonstration, while the targeted demonstrations $D_\mathrm{extra}$ fill in the gaps for the features that were initially underspecified. This enables recovery of reward weights that reflect the human's complete preferences, rather than only those objectives that happened to receive adequate supervision in the original demonstration set.

\section{Simulated Experiments}

We first evaluate our method with simulated human feedback. In simulation, we demonstrate that targeted querying for additional demonstrations on predicted underspecified features improves reward learning. We also show that our framework scales to settings with extraneous, task-irrelevant features.

\subsection{Experimental Setup}

\begin{figure}
    \centering
    \includegraphics[width=\linewidth]{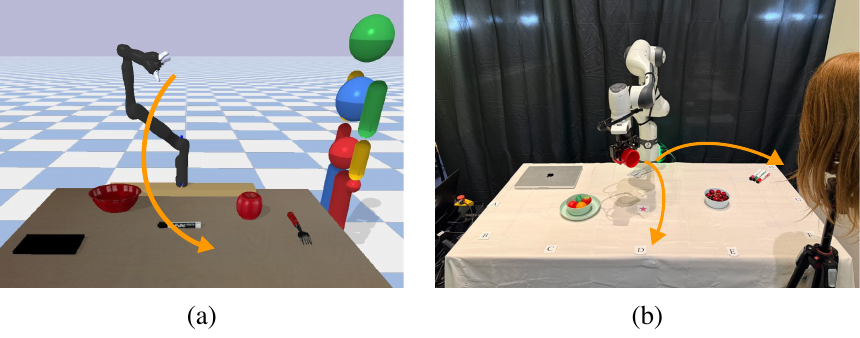}
    \vspace{-.5cm}
    \caption{Experimental environments. (a) JacoRobot simulated environment, where a 7-DoF Jaco arm manipulates a coffee cup across a cluttered tabletop. (b) User study setup with a real Franka Emika FR3 robot arm in a corresponding tabletop manipulation task.}

    \label{fig:environments}
\end{figure}

\begin{figure*}
    \centering
    \includegraphics[width=\linewidth]{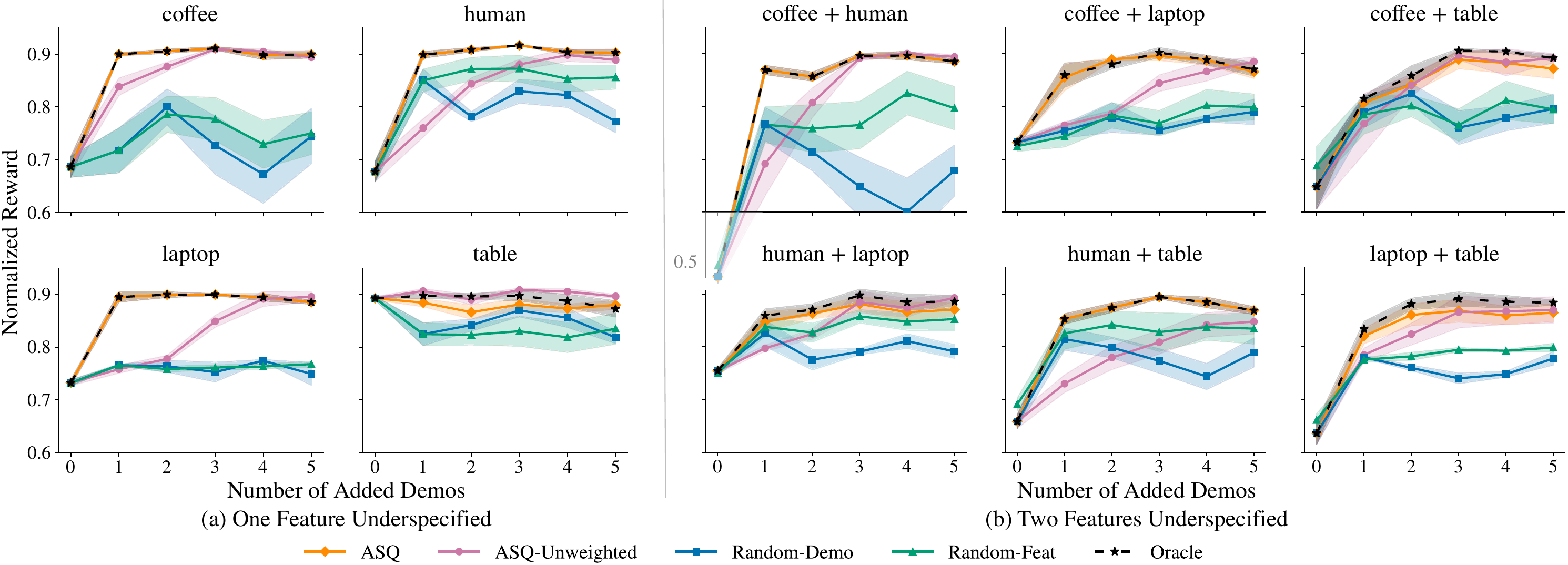}
    \caption{Reward recovery in \textit{JacoRobot} when the initial demonstrations underspecify (a) a single feature or (b) two features. ASQ consistently outperforms the random baselines and approaches Oracle performance, recovering the true reward with fewer targeted demonstrations. Lines show mean normalized reward across 5 seeds and shaded regions denote standard error.
    }
    \label{fig:jaco_base}
\end{figure*}

We evaluate our approach in a continuous robotic manipulation task where a 7-DoF Jaco arm carries a coffee cup across a cluttered tabletop. This \textit{JacoRobot} environment consists of a robot arm manipulating a coffee cup using the PyBullet simulator (\autoref{fig:environments}a). The environment includes several objects positioned on or near the table: a laptop, human, apple, bowl, fork, and marker. We randomly sample start-goal pairs and smoothly perturb the shortest-path trajectories. Trajectories consist of 21 waypoints, each with 109 dimensions: the robot's joint configuration, $xyz$ positions of all joints and objects, and their rotation matrices. We define four primary features: 1) \textit{laptop}: $xy$-distance of the end-effector to the laptop, 2) \textit{table}: $z$-distance of the end-effector above the table, 3) \textit{human}: $xy$-distance of the end-effector to the human, and 4) \textit{coffee}: length of the end effector's up axis projected onto the world up axis, a measure of how upright the coffee cup is.  The robot should avoid the laptop and human, remain close to the table, and ensure that the coffee cup does not spill. For experiments with extraneous objects, we additionally define distractor features: $xy$-distances to the 5) apple, 6) bowl, 7) fork and 8) marker.

We generate 50000 trajectories across 5000 random start-goal pairs (10 randomly perturbed trajectories per pair). The trajectory pool is split into 60\% training, 20\% validation, and 20\% test sets. 
Simulated human demonstrators are modeled as Boltzmann-rational agents with ground-truth preference weights $\theta^*$ and per-feature rationality coefficients $\beta$. Setting $\beta_i$ to a high value produces demonstrations that reliably optimize $\phi_i$, while setting $\beta_i$ to zero simulates underspecification. We note that this is a conservative model of human behavior; real demonstrators may de-prioritize rather than entirely ignore non-targeted features.

\smallskip
\noindent\textbf{Methods.} We compare five approaches. 1) \textbf{ASQ} uses variance-based detection to identify the most likely underspecified features, requests demonstrations targeting those features, and applies demonstration-specific weights during reward learning. 2) \textbf{ASQ-Unweighted} ablates the rationality weighting, applying uniform weights across all demonstrations, isolating the contribution of the weighting scheme from the querying strategy. 3) \textbf{Random-Demo} requests each additional demonstration for $N$ randomly selected features. 4) \textbf{Random-Feat} selects $N$ random features at the start, and then all additional demonstrations target those same features. 5) \textbf{Oracle} uses the ground-truth underspecified features instead of the variance-based predictions. In all conditions, the simulated human provides 10 initial demonstrations with $N$ features underspecified, followed by zero to five additional targeted demonstrations. 

For evaluation, we select the top-$k$ trajectories from the test set with the highest learned reward, compute their average ground-truth reward, and normalize by the range of ground-truth rewards across all test trajectories.
We use $k=20$ for all experiments. All experiments are repeated across 5 seeds, and we report mean normalized reward with standard error.

\smallskip
\noindent\textbf{LLM-Based Filtering.} Real-world robotic environments often contain features that are irrelevant to the current task. To this end, we expand the JacoRobot feature set to include four additional objects. The four original features $\phi_{\text{task}}$ (laptop, table, human, coffee) remain task-relevant, while the four new features $\phi_{\text{extra}}$ (apple, bowl, fork, and marker) serve as distractors. In this 8-feature setting, we query a large language model (GPT-5.2) with the task description and the full feature list, asking it to identify which features a typical human would consider necessary for safe and successful task completion. The LLM returns a binary relevance judgment for each feature, providing a filtered feature set $\phi \subseteq \phi_{\text{task}} \cup \phi_{\text{extra}}$. Features judged as irrelevant are excluded from subsequent detection, reducing the hypothesis space and preventing false positives on distractor features. To strengthen the random baselines, we add the following constraint. If all of the randomly chosen features are from $\phi_{\text{extra}}$, then the additional demonstrations reflect the suboptimal demonstrator's initial behavior. We assume that, if a human was asked to provide demonstrations about only irrelevant features, that they would repeat their original behavior rather than optimize for features for which they have no preferences.

\subsection{Results}

\begin{figure*}
    \centering
    \includegraphics[width=\linewidth]{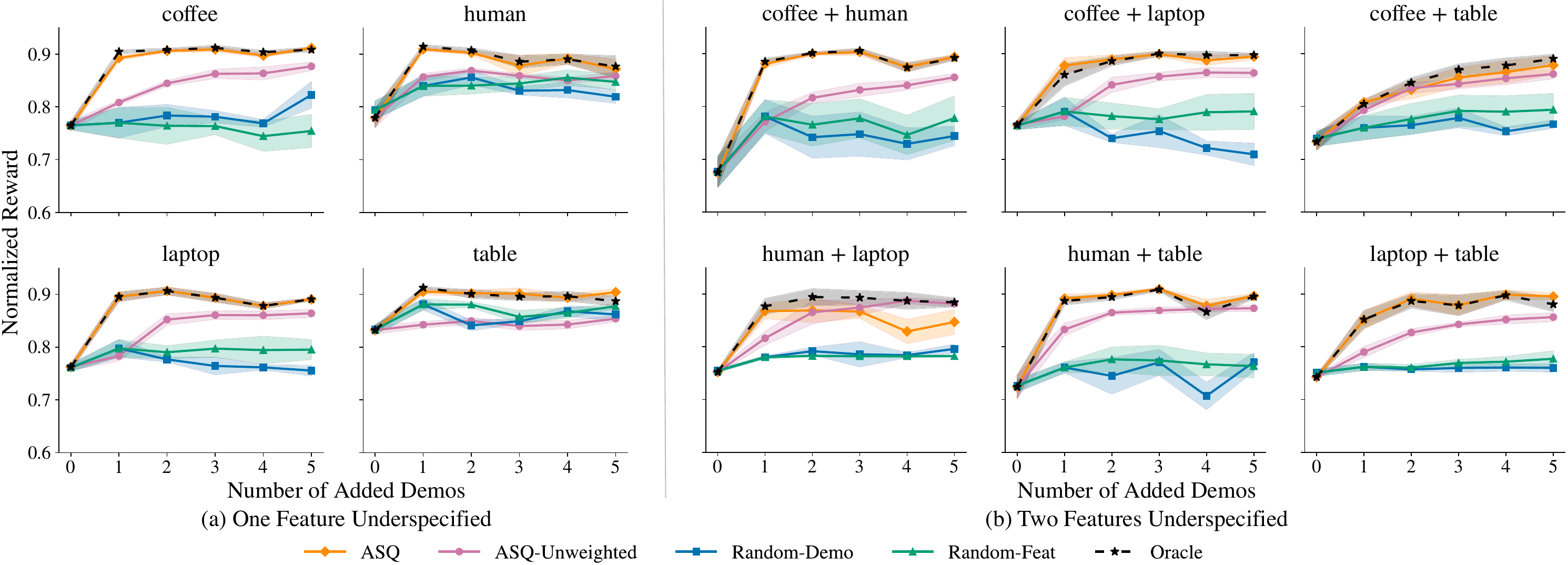}
    \caption{Reward recovery in \textit{JacoRobot} in the 8-feature setting that augments the feature set of 4 task-relevant features with 4 irrelevant distractor objects, with (a) one task-relevant feature underspecified and (b) two task-relevant features underspecified. LLM-based filtering prunes the distractors before variance-based detection, allowing ASQ to focus queries on task-relevant features. ASQ matches or exceeds the random baselines across all single-feature and pairwise comparison conditions, demonstrating robustness to extraneous environmental features. Lines show mean normalized reward across 5 seeds and shaded regions denote standard error.
    }
    \label{fig:jaco_extra}
\end{figure*}

\noindent\textbf{Task-Relevant Features.} We first evaluate settings where all features in the environment are task-relevant, eliminating the need for LLM-based filtering. The first condition that we consider is where exactly one feature is underspecified during initial demonstrations ($\beta_i=0$ for underspecified $\phi_i$). We simulate a human providing targeted corrective demonstrations for the requested features $\phi_{\mathrm{under}}$ by setting $\beta_j=0$ for all $\phi_j \notin \phi_{\mathrm{under}}$. In the real world, it is possible that a human demonstrator may actually provide more informative targeted demonstrations. Specifically, a real-life demonstrator may instead just de-prioritize, rather than ignore, the features optimized for in the initial demonstrations. As shown in \autoref{fig:jaco_base}a, ASQ outperforms the random baselines in all four conditions. When coffee, human, or laptop is underspecified, ASQ recovers reward more quickly and achieves higher final performance. The table feature, however, shows that underspecification is not synonymous with suboptimality. When a demonstrator attends to three features, the resulting trajectories could happen to also perform well on the fourth feature, leading to a high normalized reward at the start. Secondly, we evaluate all six pairwise combinations of two underspecified features. We show in \autoref{fig:jaco_base}b that ASQ's advantage persists when two features are simultaneously underspecified, although the gain over the baselines varies depending upon the feature combination. 

\smallskip
\noindent\textbf{Scaling to Extraneous Features.} In the 8-feature setting, we evaluate whether LLM-based filtering enables our method to maintain robust performance despite the presence of task-irrelevant distractors. For these experiments, we simulate the demonstrator to have $\theta_i=0$ for $\phi_i \in \phi_{\text{extra}}$, reflecting that the user's preferences assign no weight to task-irrelevant features. We show in \autoref{fig:jaco_extra}a that ASQ outperforms the baselines in the single-feature setting. Compared to the baselines, ASQ improves reward learning more rapidly in the early demonstrations, reducing the number of corrective demonstrations required from the demonstrator for reward recovery. When two features are underspecified, ASQ also shows an advantage for all pairwise combinations (\autoref{fig:jaco_extra}b). These results demonstrate the effectiveness of the method in settings where robots must distinguish between meaningful task features and irrelevant environmental distractors.

\section{User Study}\label{sec:userstudy}

To evaluate whether feature-based explanations help humans provide more informative corrective demonstrations, we conduct a user study comparing our explanation-guided approach against natural alternatives.

\subsection{Experiment Design}

Our primary investigation concerns whether explanations that identify underspecified features in interpretable terms lead to better corrective demonstrations than simply observing the robot's behavior. While simulation experiments can evaluate whether targeted demonstrations improve reward learning, they cannot capture whether real users can translate different forms of feedback into actionable guidance. A robot might convey its limitations by rolling out its current policy, allowing the user to observe failures directly. Alternatively, it might provide an explicit explanation naming the features it misunderstands. We hypothesize that the latter produces demonstrations more useful for reward recovery. Rather than leaving users to diagnose the problem themselves, explanations direct attention to the specific features requiring supervision.

We conducted a user study under an IRB-approved protocol using a real Franka Emika FR3 robot arm in a tabletop setting. The setup, shown in \autoref{fig:environments}b, preserved the manipulation task from our simulation experiments while grounding the interaction in a realistic physical context. The study followed a within-subjects design that varied feedback requests at three levels. 1) \textbf{Unguided}: the robot requested demonstrations with no additional context. 2) \textbf{Rollout}: the robot first executed a trajectory according to its currently learned reward function, then requested demonstrations. 3) \textbf{Explanation}: the robot provided a natural language explanation identifying the feature it was uncertain about, then requested demonstrations. The explanation followed a fixed template that named the underspecified feature directly. 

Participants were told all four task objectives upfront (keeping the mug upright, close to the table, away from the laptop, and away from people) and reminded of them before each condition (Appendix \ref{app:user-study-details}). Ensuring participants cared about all task-relevant features isolates the effect of attentional guidance on demonstration quality, letting us investigate which querying method best directs attention to the underspecified dimension and yields more informative corrective data. Each participant completed all three conditions. We counterbalanced condition order by assigning participants evenly across the six possible orderings. Within each condition, participants first completed a familiarization phase to become comfortable with the interaction modality, followed by two experimental tasks. The familiarization phase used \textit{coffee} as the underspecified feature: participants stood in the demonstrator position while the robot queried according to the assigned condition, after which they provided three practice demonstrations. The first experimental task targeted \textit{human} as the underspecified feature, while the second task used \textit{laptop} as the underspecified feature. For each task, participants provided three demonstrations. After completing both tasks within a condition, participants filled out a survey about their experience with that method before proceeding to the next condition. A final comparative survey collected overall preferences across methods.

\smallskip
\noindent\textbf{Manipulated Variables.} We compare the type of feedback provided before demonstration collection: no feedback (\textit{Unguided}), behavioral rollout (\textit{Rollout}), or feature-based explanation (\textit{Explanation}). The explanation always correctly identified the underspecified feature, enabling us to evaluate whether this form of communication helps users provide better demonstrations independent of prediction accuracy.

\smallskip
\noindent\textbf{Hypotheses.} We test three hypotheses regarding the benefits of feature-based explanations for eliciting corrective demonstrations: \textbf{H1.} Providing an explanation when querying for demonstrations leads humans to provide corrective demonstrations that yield better aligned rewards. \textbf{H2.} Explanations reduce the cognitive load of providing a corrective demonstration relative to conditions where users must base their correction on observed behavior. \textbf{H3.} Participants will subjectively prefer the \textit{Explanation} condition over the alternative methods.

\smallskip
\noindent\textbf{Dependent Measures.} We evaluate each condition using both objective and subjective metrics. For objective evaluation, we measure normalized reward recovery: for each participant, condition, and task, we train a reward function using MaxEnt IRL on their collected demonstrations combined with a fixed set of initial demonstrations, then compute the normalized ground truth reward of trajectories selected by the learned reward function. 
For subjective evaluation, we administer the NASA Task Load Index (NASA-TLX)~\cite{hart1988development} questionnaire measuring cognitive workload after each condition, as well as a forced-choice preference question in the final comparative survey, asking participants to select which method they preferred overall.

\smallskip
\noindent\textbf{Participants.}  We recruited $N = 12$ participants (6 female, 6 male; age $M = 28.5$, $SD = 12.5$) with moderate prior experience controlling or operating robots ($M = 4.2$, $SD = 1.5$ on a 1--7 scale).

\subsection{Analysis}

\begin{figure}[t]
    \centering
    \includegraphics[width=\linewidth]{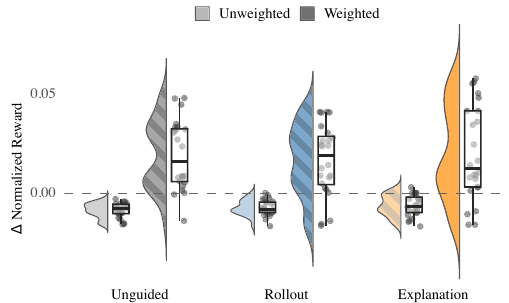}
    \caption{Change in normalized reward by condition. Hashed indicates an ablative configuration; applying weighting to demonstrations that couldn't have been known to be targeted, or not weighting demonstrations that can be presumed to be targeted. Only the \textit{Explanation} condition yielded positive improvement on average. To account for stochasticity in the process, reward learning was repeated 100 times and the median performance used. Since participants demonstrated two tasks, the average of the change in normalized reward is used.
    }      
    \label{fig:reward_improvement}
\end{figure}

 All analyses use linear mixed-effects models with participant as a random effect and condition order as a covariate to account for the within-subjects design.

  \smallskip
\noindent\textbf{H1: Reward Improvement} was supported. We found a significant main effect of condition on normalized reward improvement, $F(2,22) = 14.94$, $p < .001$. Estimated marginal means revealed that the \textit{Explanation} condition ($M = 0.021$, $SE = 0.005$) yielded significantly greater reward improvement than both \textit{Unguided} ($M = -0.018$, $SE = 0.005$; $t = -6.31$, $p < .001$) and \textit{Rollout} ($M = -0.016$, $SE = 0.005$; $t = -6.06$, $p < .001$). No difference was observed between \textit{Unguided} and \textit{Rollout} ($p = .966$). Demonstrations elicited via feature-based explanations thus produce better reward recovery when combined with demonstration-specific rationality weighting. This weighting presumes that the human attended to the named features, an assumption that is only justified in \textit{Explanation}, where we directed the demonstrator's attention to those features. In contrast, in \textit{Unguided} and \textit{Rollout}, the human may have chosen to focus on any of the four task-relevant features, so applying the same weighting scheme would attribute attention we cannot verify.

\begin{figure}[t]
    \centering
    \includegraphics[width=\linewidth]{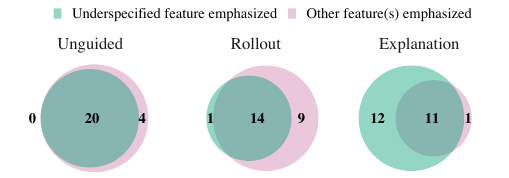}
    \caption{Whether participants reported emphasizing the underspecified feature in their demonstrations, emphasizing other features, or both. Each of the twelve participants demonstrated two tasks in each condition.
    }      
    \label{fig:feature_emphasis}
\end{figure}

To separate the effect of the explanation from the weighting it enables, we compared all-weighted and all-unweighted reward variants. Within \textit{Explanation}, all-weighted rewards ($M = 0.021$, $SE = 0.007$) were higher than all-unweighted rewards ($M = -0.017$, $SE = 0.003$), $t(23) = 4.56$, $p < .001$, mean difference $= 0.038$. As shown by the hashed entries of \autoref{fig:reward_improvement}, assuming correct weighting substantially narrows the gap between \textit{Explanation} and the other conditions. Moreover, when only all-weighted rewards were analyzed, condition differences were not significant (all $p \geq .183$), leaving only a directional trend favoring \textit{Explanation}. However, the demonstrations themselves did change meaningfully across conditions. Per-participant feature analysis (Appendix~\ref{sec:userstudy_data}) shows that under \textit{Explanation}, variance on the features the robot did not flag widens relative to the other conditions, consistent with participants relaxing their attention on objectives that were not identified as uncertain. Participants' self-reports corroborate this shift: in the \textit{Explanation} condition, participants overwhelmingly reported emphasizing the named feature in their demonstrations, whereas in \textit{Rollout} they frequently misidentified the underspecified feature (\autoref{fig:feature_emphasis}).

  \smallskip
\noindent\textbf{H2: Cognitive Workload} was not supported. We found no significant effect of condition on NASA-TLX composite scores, $F(2,22) = 0.76$, $p = .480$. Workload was comparable across \textit{Unguided} ($M = 28.1$, $SE = 3.7$), \textit{Rollout} ($M = 28.3$, $SE = 3.7$), and \textit{Explanation} ($M = 25.6$, $SE = 3.7$), with no significant pairwise differences (all $p \geq .530$). Although we hypothesized that explanations would reduce cognitive load by narrowing what the demonstrator must consider, the physical demands of the task remained largely invariant across conditions: participants kinesthetically guided the same robot through the same motion for the same number of demonstrations. In this controlled setting, attentional guidance changes what the demonstrator thinks about, but not the physical or temporal effort each demonstration requires. Analysis of individual TLX subscales revealed no significant effects for any dimension (all $p > .19$).  Demonstration duration also did not differ significantly across conditions, $F(2,22) = 1.25$, $p = .307$ (Unguided: $M = 17.3$s; Rollout: $M = 14.6$s; Explanation: $M = 16.6$s). In a practical deployment, we hypothesize that the workload benefit of targeted querying would likely manifest as a reduction in the number of demonstrations required to correct the misaligned reward. Investigating whether explanation-guided querying reduces total teaching burden in less constrained settings, where the human controls when to stop providing demonstrations, is a direction for future work.

  \smallskip
\noindent\textbf{H3: Subjective Preference} was not supported. Five of twelve participants preferred \textit{Explanation}, five preferred \textit{Rollout}, and the remaining two preferred \textit{Unguided} (exact multinomial vs.\ uniform: $\chi^2 = 1.50$, $p = .616$). While preferences did not significantly favor any condition, the qualitative reasons participants gave were informative. Participants who preferred \textit{Rollout} often described the rollout as a concrete behavior to repair rather than as an abstract representation of the robot's misunderstanding. P6 explained that it felt easier to teach the robot by providing the ``smallest perturbation to an existing trajectory,'' while P7 noted that ``seeing the path the robot thought was best'' made it easier to identify and correct flaws. Participants preferring \textit{Explanation} emphasized explicit guidance and reduced cognitive load. P4 noted that ``it was nice to have specific direction about where to target the efforts of my demonstration,'' and P8 explained that being told ``exactly what [the robot] was unsure of'' made it easier to focus on one aspect of the task, rather than ``trying to do several things at once.'' Those preferring \textit{Unguided} cited varied reasons, including task familiarity and freedom to give natural demonstrations. P11 noted that the condition allowed them to offer the ``easiest demonstrations,'' whereas knowing exactly what the robot was confused about required them to ``think a bit more'' about how to convey the correction. These responses suggest that no single feedback modality works for all users.

\section{Conclusion}

We introduced ASQ, a framework that addresses a fundamental limitation of learning from demonstrations: from demonstrations alone, a robot cannot determine whether feature indifference reflects the demonstrator's true preferences or simply indicates incomplete supervision. Our insight is that optimization leaves a statistical signal in feature variability, enabling  identification of underspecified features through Bayesian model selection over observed variance patterns. By generating natural language queries that direct the demonstrator's attention to these specific dimensions, ASQ resolves reward ambiguity with minimal human effort. In simulation, ASQ improves reward learning using only a small number of targeted demonstrations in a continuous manipulation domain. Our user study with a physical robot confirmed that our approach translates to real human demonstrators, showing that feature-based explanations produce more aligned reward functions compared to alternative modalities. 

\smallskip
\noindent\textbf{Limitations and Future Work.} Although ASQ successfully resolves reward ambiguity, several limitations remain. Our approach treats feature attention as binary and static across a demonstration session, but in practice, attention may fluctuate within a single trajectory. 
We also assume conditional independence of feature variances given optimization status, yet features are often correlated in practice.
Beyond these modeling choices, our framework elicits feedback exclusively through kinesthetic demonstrations, but humans communicate preferences through diverse modalities. Augmenting the framework to incorporate natural language feedback from the user or comparative judgments could resolve ambiguity more efficiently. In our method, features the user is indifferent to are likely task-irrelevant and will not be queried if detected by LLM-based filtering. Nonetheless, a natural extension is a two-stage interaction in which the robot first names uncertain features and asks the user whether they matter before collecting demonstrations. Additionally, our user study involved a modest sample size, and larger-scale studies across diverse populations would strengthen confidence in our findings. Our experiments also focused on a manipulation domain with a fixed set of objects. Evaluating ASQ in more complex or dynamic environments where objects appear, disappear, or move throughout the task would further validate the robustness and broader applicability of our method.

\section*{Acknowledgments}

This work was funded by a seed grant from the MIT Schwarzman College of Computing's  Social and Ethical Responsibilities of Computing program.

\vspace{1em}

\bibliographystyle{plainnat}
\bibliography{references}

\clearpage
\appendix

\subsection{Implementation Details}
\smallskip
\noindent\textbf{Reference Variance Distributions.} To construct the reference distributions $P(\sigma_i^2 \mid o_i=0, \phi_i)$ and $P(\sigma_i^2 \mid o_i=1, \phi_i)$ for each feature $\phi_i \in \phi$, we generate demonstration pools under all possible optimization configurations. Specifically, we enumerate all subsets of $\phi$. For each subset $\mathcal{S} \subseteq \phi$, we instantiate a Boltzmann-rational demonstrator with preference weights $\beta_i = 80$ for \textit{JacoRobot} or $\beta_i = 50$ for \textit{GridRobot} if $\phi_i \in \mathcal{S}$ and $\beta_i=0$ otherwise, and generate 20 demonstrations. For each feature $\phi_i$, we then partition those demonstration pools based on whether $\phi_i$ was being optimized: pools where $\phi_i \in \mathcal{S}$ contribute to the optimized distribution, while pools where $\phi_i \notin \mathcal{S}$ contribute to the non-optimized distribution. To build empirical variance distributions, we repeatedly random sample 10 demonstrations from each pool and compute the sample variance of $\phi_i$ across the sample. This sampling procedure is repeated 500 times, yielding empirical distributions that we fit with chi-squared distributions.

\smallskip
\noindent\textbf{Training Parameters. }MaxEnt IRL training proceeds for 200 iterations in the \textit{GridRobot} domain and 1000 iterations in the \textit{JacoRobot} domain, using Adam optimization with learning rates of 0.01 and 0.001, respectively.

\subsection{LLM Prompt for Feature Importance}

In the JacoRobot environment with extraneous objects, we use GPT-5.2 with high reasoning effort to evaluate whether each feature is necessary for sensible robot behavior. The model is queried once per feature using structured JSON output.

\begin{promptbox}
\setlength{\parindent}{1.5em}

\noindent The following is an instruction for a task that the agent needs to complete. instruction = A 7-DoF Jaco robot arm is carrying a coffee cup above a tabletop environment. The arm is carrying a fragile ceramic coffee cup that is at risk of slipping. Various objects are on the table. A human is right next to the table.

The current available feature names are:
laptop (distance from end-effector to laptop)
table (height of end-effector above table surface)
human (distance from end-effector to human)
coffee (coffee cup orientation (tilt from upright))
apple (distance from end-effector to apple)
bowl (distance from end-effector to bowl)
fork (distance from end-effector to fork)
marker (distance from end-effector to marker)

We are evaluating whether \{feature\} (\{feature\_description\}) is part of the set of features that a typical human would consider necessary for this task.

Please return either a 1 or 0 for the following feature:
feature = \{feature\} (\{feature\_description\}).

A feature should be labeled 1 only if a typical human would expect that ignoring it would commonly lead to a clearly undesirable outcome in this 
specific task. If it is unclear whether the feature would meaningfully affect 
behavior in this context, then it should be labeled 0.

Please reason step by step internally.

Provide your answer as ONLY JSON in one of these two forms:
\{"label": 0\} or \{"label": 1\} Do not include any surrounding explanation or text.\end{promptbox}

\subsection{GridRobot Navigation}

\begin{wrapfigure}{r}{0.2\textwidth}
    \centering
    \includegraphics[width=0.2\textwidth]{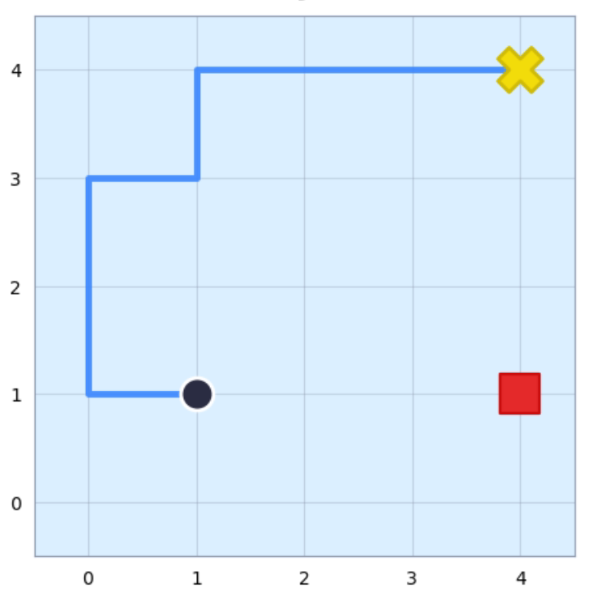}
    \caption{GridRobot navigation simulated environment.}
    \label{fig:gridrobot_env}
\end{wrapfigure}

\begin{figure}[h]
    \centering
    \includegraphics[width=\linewidth]{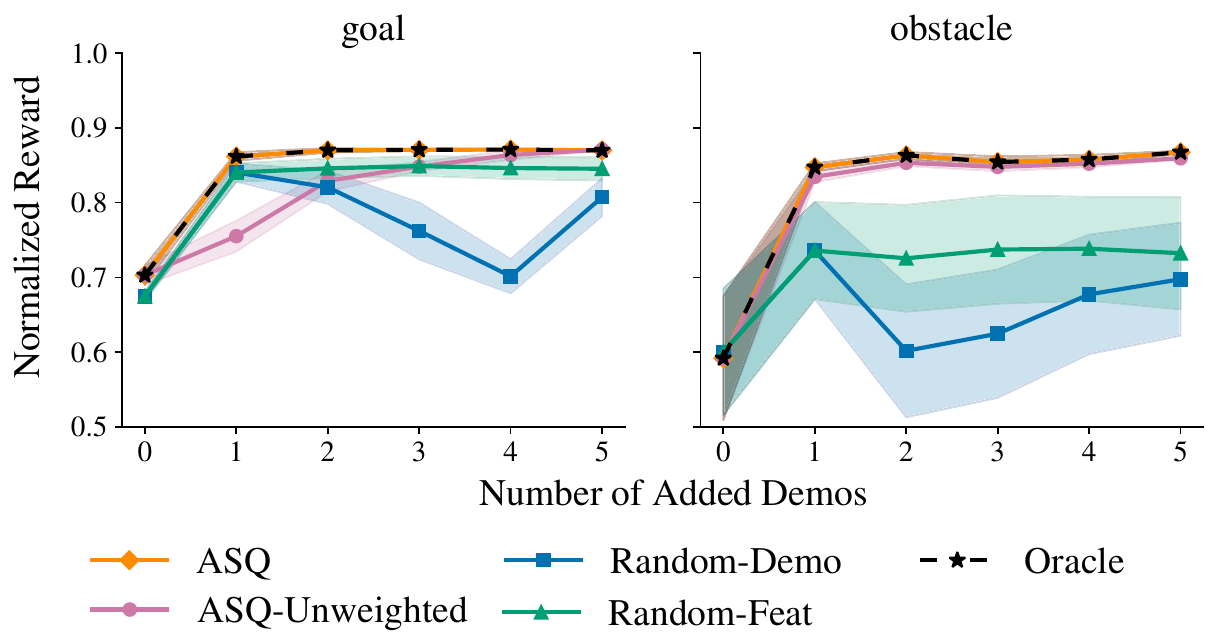}
    \caption{GridRobot where the initial demonstrations have one underspecified feature.}
    \label{fig:grid_base}
\end{figure}

Along with the JacoRobot environment, we evaluate our approach in the GridRobot environment: a simulated discrete 2D navigation task. In the GridRobot domain, an agent must navigate from a start position to a goal position on a 5x5 grid while avoiding an obstacle (\autoref{fig:gridrobot_env}). Trajectories are sequences of 9 states, resulting in an 18-dimensional input comprised of the $x$ and $y$ coordinates of each state. We define two features: 1) \textit{goal}: distance to goal, and 2) \textit{obstacle}: distance to obstacle. The agent should maximize distance from the obstacle while minimizing distance to the goal. We generate 30000 trajectories across 3000 random start-goal-object configurations (10 trajectories per configuration), and split the trajectory pool into 60\% training, 20\% validation, and 20\% test sets. We show that ASQ matches oracle performance while the baselines plateu at lower values. (\autoref{fig:grid_base}).

\subsection{User Study Details}
\label{app:user-study-details}

\noindent\textbf{Participant Instructions.} After providing informed consent, participants were introduced to the robot and practiced kinesthetic guidance. They were told that the robot was learning to carry a mug of coffee across the table and that it was ``bad at this task.'' Four task objectives were stated explicitly: (1) keeping the mug upright, (2) staying close to the table, (3) staying away from electronics, and (4) staying away from people. Participants were instructed that they would teach the robot in three modes (our experimental conditions) and that their job was to ``provide the most specific demonstration that you possibly can that explicitly focuses on just what the robot doesn't understand.'' These objectives were repeated before interacting with each condition.

Each of the three conditions consisted of a familiarization phase followed by two experimental tasks. The familiarization phase used \textit{coffee} as the underspecified feature, while the two experimental tasks targeted \textit{human} and \textit{laptop} as the underspecified features. During the familiarization round, each condition was introduced with the following language:

\begin{itemize}
\item \textit{Unguided}: ``The robot will ask you to show it how to move the mug to the edge of the table. It won't say or show you anything. Choose any target, from the positions A through H, that allows you to best demonstrate the task.''

\item \textit{Rollout}: ``The robot will show you how it thinks it should move, and then ask you to demonstrate. You should focus your demonstration on just what it seems the robot doesn't understand. Choose any target, from the positions A through H, that allows you to best demonstrate to the robot just what it doesn't understand.''

\item \textit{Explanation}: ``The robot will provide an explanation that specifies the part of the task that it doesn't understand. You should focus your demonstration on just what the robot says that it doesn't understand. Choose any target, from the positions A through H, that allows you to best demonstrate to the robot what it doesn't understand.''
\end{itemize}

All conditions involved language prompts read to the human by the experimenter. The prompts were:

\begin{itemize}
\item \textit{Unguided} (generic prompt): ``Show me how to move the mug to the edge of the table.''

\item \textit{Rollout} (generic prompt with behavioral context): The robot said ``This is how I think I should do it,'' executed a trajectory according to its currently learned reward, and then said ``Show me how best to move to the edge of the table.''

\item \textit{Explanation} (feature-specific prompt): ``I am uncertain how to handle \texttt{[feature]}. Show me how best to move to the edge of the table, focusing on \texttt{[feature]},'' where \texttt{[feature]} was replaced with the underspecified feature for the task.
\end{itemize}

While the presence of language was controlled, both the amount and specificity of language varied across conditions.

\smallskip
\noindent\textbf{Statistical Methodology.} All inferential analyses were conducted in R using linear mixed-effects models fit with restricted maximum likelihood (REML) via the \texttt{lme4} package~\cite{bates2015fitting}. Each model included condition (three levels: \textit{Unguided}, \textit{Rollout}, \textit{Explanation}) and condition order (position 1--3) as fixed effects and participant as a random intercept, e.g., \texttt{outcome $\sim$ condition + position + (1\,|\,participant)}. This specification assumes (i) normally distributed residuals with constant variance across conditions, (ii) normally distributed random intercepts, and (iii) independence of observations across participants. Denominator degrees of freedom and $p$-values for fixed effects were obtained via the Satterthwaite approximation provided by the \texttt{lmerTest} package~\cite{kznetsova2017lmertest}. Omnibus $F$-tests for the condition factor were supplemented by estimated marginal means and pairwise contrasts computed with the \texttt{emmeans} package~\cite{lenth2024emmeans}, with Tukey's HSD adjustment applied to all pairwise comparisons to control the family-wise error rate within each outcome. No additional correction was applied across the three primary outcomes (reward improvement, TLX composite, and preference), as each addresses a distinct hypothesis.

\smallskip
\noindent\textbf{Workload Discussion.} We hypothesized two mechanisms by which demonstrator guidance might reduce demonstrator workload: 1) focusing on specific aspects of the task (e.g. staying away from the laptop) would be less physically demanding than addressing all of them (e.g. staying away from the laptop while keeping the cup upright); and 2) inferring the underlying issue from a single rollout would be challenging and operators would rate their perceived success as lower.

Our analysis indicated there were no statistically significant differences between conditions. Generally, the workload of the task was low, perhaps because there was no time pressure or explicit performance feedback integrated into the study. There was a trend that workload was lower for the Rollout condition, visible largely in the Perceived Failure and Effort subscales. Participants' preferences and open-ended feedback did not surface a theory that might explain the observed trend. Additional experiments measuring workload under more strenuous task conditions would be necessary to detect possible effects.

\subsection{User Study Data}
\label{sec:userstudy_data}

\begin{figure*}
    \centering
    \includegraphics[width=0.75\linewidth]{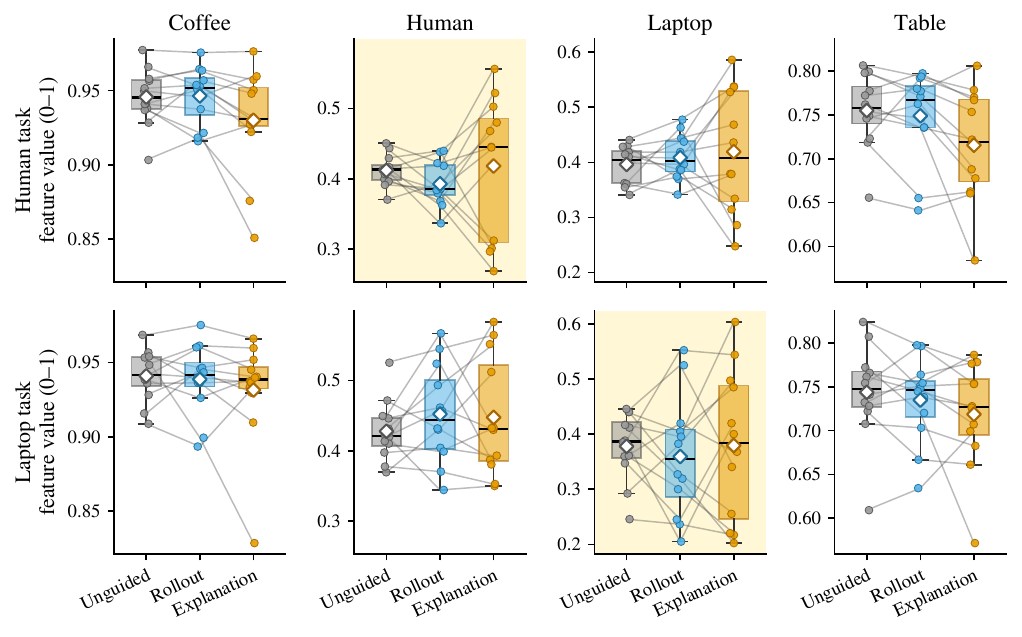}
    \caption{Per-participant demonstration feature values across the three user-study conditions. Each column shows one task-relevant feature; rows correspond to the two experimental tasks, and the highlighted subplots mark the feature that the \textit{Explanation} condition explicitly named as underspecified. Higher feature values correspond to behavior better aligned with the objective. Boxes show the median and IQR, white diamonds mark group means, individual dots mark per-participant means, and the gray lines connect each participant's three repeated measures across conditions.}
    \label{fig:user_study_data}
\end{figure*}

We examined whether the three feedback conditions led participants to produce behaviorally distinct demonstrations. We computed the four task-relevant feature values for every collected demonstration and aggregated to per-participant means within each condition and task. Figure~\ref{fig:user_study_data} shows the resulting distributions. Under the equal weight ground-truth reward $\theta^*_i = 1$ for each task-relevant feature $\phi_i$, higher feature values correspond to demonstrations that better reflect the desired behavior.

While per-participant means on the underspecified feature in each task are largely similar across all three conditions (human task: \textit{Unguided} $0.412$, \textit{Rollout} $0.393$, \textit{Explanation} $0.418$; laptop task: \textit{Unguided} $0.378$, \textit{Rollout} $0.360$, \textit{Explanation} $0.380$), the \textit{Explanation} mean is slightly higher than the other conditions in each task. Higher values correspond to demonstrations that, on average, kept the end-effector farther from the named object. This is directionally consistent with the intended effect of \textit{Explanation}, although the magnitude is small. Moreover, visible behavioral changes appear on the other features. In the human task, the laptop, table, and coffee features all show substantial standard-deviation increases under \textit{Explanation}: $3.36\times$ and $2.76\times$ for laptop (relative to \textit{Unguided} and \textit{Rollout}), $1.51\times$ and $1.23\times$ for table, and $1.85\times$ and $1.83\times$ for coffee. In the laptop task, the human, table, and coffee features show similar widening ($1.98\times$ and $1.21\times$, $1.12\times$ and $1.27\times$, $2.06\times$ and $1.50\times$, respectively). This pattern suggests participants were following the instructions to focus on the underspecified feature. When participants were asked to give a demonstration that explicitly focused on the feature the robot did not understand, their behavior on the other features became correspondingly less constrained.

\begin{figure*}
    \centering
    \includegraphics[width=\linewidth]{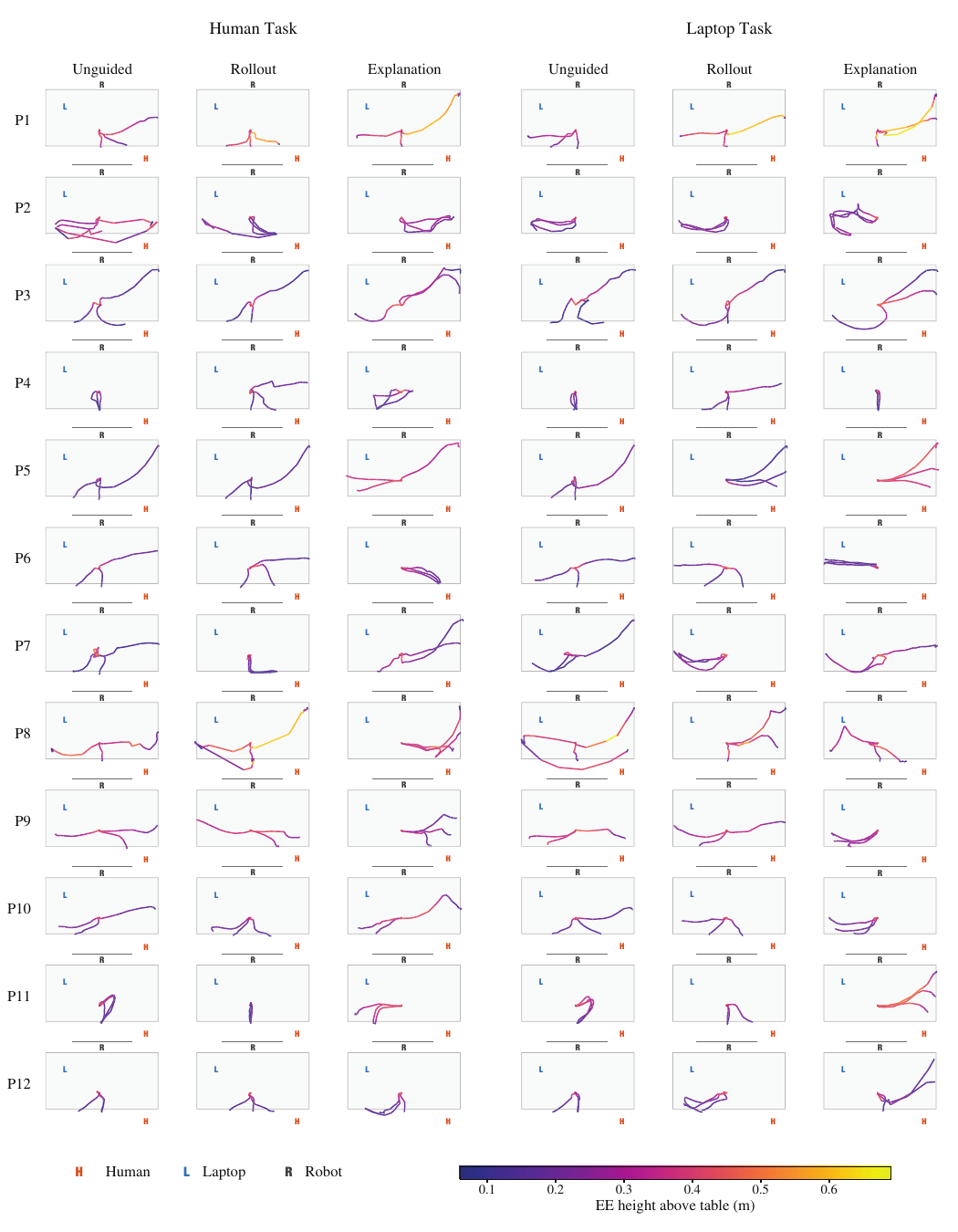}
    \caption{End-effector trajectories from all user study demonstrations. Each panel shows the three demonstrations provided by one participant (rows, P1-P12) under one condition (\textit{Unguided}, \textit{Rollout}, \textit{Explanation}) on one task (\textit{human}, \textit{laptop}).}
    \label{fig:user-study-demos}
\end{figure*}

Additionally, standard deviations on the underspecified feature itself widen sharply under \textit{Explanation}: in the human task, from $0.022$ (\textit{Unguided}) and $0.031$ (\textit{Rollout}) to $0.099$, and in the laptop task, from $0.062$ (\textit{Unguided}) and $0.109$ (\textit{Rollout}) to $0.137$. This result reveals a limitation of the study. The prompt ``the robot is uncertain about distance to the laptop'' admits two reasonable interpretations: some participants understood it as a request to demonstrate maintaining a safe working distance, and consequently ended up \emph{closer} to the laptop (lower values) than they had under either \textit{Unguided} or \textit{Rollout}; others understood it as a request to keep the end-effector as \emph{far} from the laptop as possible, producing trajectories with greater clearance (higher values). Together, these interpretations push the demonstrations in opposing directions. As a result, a natural extension for future work is to pair the feature name with visualizations of the intended behavior. The full set of 216 demonstrations across all participants, conditions, and tasks is shown in Figure~\ref{fig:user-study-demos}.

\subsection{Contribution Statement}

HM led the simulation experiments and associated system implementation. NW led the design and analysis of the user study. Both authors conducted the user study and contributed to overall system development. AB supervised the project and secured funding. All authors contributed to the conceptualization and writing.

\end{document}